\documentclass[sigconf]{acmart}
\AtBeginDocument{%
  \providecommand\BibTeX{{%
    \normalfont B\kern-0.5em{\scshape i\kern-0.25em b}\kern-0.8em\TeX}}}

\setcopyright{rightsretained}
\copyrightyear{2023}
\acmYear{2023}
\acmConference[WSDM '23]{Proceedings of the Sixteenth ACM International Conference on Web Search and Data Mining}{February 27-March 3, 2023}{Singapore, Singapore}
\acmBooktitle{Proceedings of the Sixteenth ACM International Conference on Web Search and Data Mining (WSDM '23), February 27-March 3, 2023, Singapore, Singapore}
\acmDOI{10.1145/3539597.3570415}
\acmISBN{978-1-4503-9407-9/23/02}
\acmPrice{15.00}
\acmISBN{978-1-4503-9407-9/23/02}

\usepackage{times}
\usepackage{latexsym}
\usepackage{graphicx} %  added
\usepackage{amsfonts} % added
\usepackage{multirow}
\usepackage{booktabs}
\usepackage{ulem}
\usepackage{hyperref}

\begin{document}

%\title{Get the Point! Graph Enhanced Candidate Retrieval for Zero-shot Entity Linking}
%\title{GER: Graph Enhanced Representation for Candidate Retrieval in Zero-shot Entity Linking}
%\title{Improving Zero-shot Entity Retrieval via Graph based Word-level Information}
%\title{Learning Fine-Grained Graph-based Word-level information for Zero-shot Entity Retrieval}
%\title{Modeling Fine-grained Information via Hierarchical Graph Neural Network for Zero-shot Entity Retrieval}
%\title{Modeling Fine-Grained Information  from Centralized Graph for Zero-shot Entity Retrieval}
\title{Modeling Fine-grained Information via Knowledge-aware Hierarchical Graph for Zero-shot Entity Retrieval}

\author{Taiqiang Wu}
\authornote{Equally contributions.}
\email{wtq20@mails.tsinghua.edu.cn}
\affiliation{%
  \institution{Shenzhen International Graduate School, Tsinghua University}
  \city{Shenzhen}
  \country{China}
}

\author{Xingyu Bai}
\email{bxy20@mails.tsinghua.edu.cn}
\authornotemark[1]
\affiliation{%
  \institution{Shenzhen International Graduate School, Tsinghua University}
  \city{Shenzhen}
  \country{China}
}

\author{Weigang Guo}
\email{jimwgguo@tencent.com}
\affiliation{%
  \institution{Tencent}
  \city{Shenzhen}
  \country{China}
}

\author{Weijie Liu}
\email{jagerliu@tencent.com}
\affiliation{%
  \institution{Tencent}
  \city{Shenzhen}
  \country{China}
}

\author{Siheng Li}
\email{lisiheng21@mails.tsinghua.edu.cn}
\affiliation{%
  \institution{Shenzhen International Graduate School, Tsinghua University}
  \city{Shenzhen}
  \country{China}
}

\author{Yujiu Yang}
\authornote{Corresponding author.}
\email{yang.yujiu@sz.tsinghua.edu.cn}
\affiliation{%
  \institution{
  Shenzhen International Graduate School, Tsinghua University}
  \city{Shenzhen}
  \country{China}
}

% %%
% %% By default, the full list of authors will be used in the page
% %% headers. Often, this list is too long, and will overlap
% %% other information printed in the page headers. This command allows
% %% the author to define a more concise list
% %% of authors' names for this purpose.

\begin{abstract}
Zero-shot entity retrieval, aiming to link mentions to candidate entities under the zero-shot setting, is vital for many tasks in Natural Language Processing.
Most existing methods represent mentions/entities via the sentence embeddings of corresponding context from the Pre-trained Language Model.
However, we argue that such coarse-grained sentence embeddings can not fully model the mentions/entities, especially when the attention scores towards mentions/entities are relatively low.
In this work, we propose GER, a \textbf{G}raph enhanced \textbf{E}ntity \textbf{R}etrieval framework, to capture more fine-grained information as complementary to sentence embeddings. 
We extract the knowledge units from the corresponding context and then construct a mention/entity centralized graph.
Hence, we can learn the fine-grained information about mention/entity by aggregating information from these knowledge units.
To avoid the graph bottleneck for the central mention/entity node, we construct a hierarchical graph and design a novel Hierarchical Graph Attention Network~(HGAN).
Experimental results on popular benchmarks demonstrate that our proposed GER framework performs better than previous state-of-the-art models.
% The code has been available at https://github.com/wutaiqiang/GER-WSDM2023.
%\footnote{The codes would be released in \href{https://github.com/wutaiqiang/GER-WSDM2023}{this repository}. }.
% Further analysis 
%and achieves 3.59 points improvement on the ZESHEL dataset.

%Specifically, we construct a novel  mention/entity centralized graph and design a Hierarchical Graph Attention Network~(HGAT) to capture the fine-grained word-level information.
% For the retrieval phase of the zero-shot entity linking task, BERT has been widely used to represent the mentions and the entities utilizing the sentence embeddings.
% However, the sentence embeddings obtained by BERT are dominated by the high-frequency words in the pre-training corpus, thus performing poorly especially when the mention/entity is a low-frequency word.
% To solve this issue, we propose a \textbf{G}raph enhanced \textbf{E}ntity \textbf{R}etrieval~(GER) framework, fusing word-level embedding with the sentence embedding from BERT.
% Specifically, we construct a mention/entity centralized graph and design a Hierarchical Graph Neural Network (HGNN) to capture the word-level information.
% Experimental results on the ZESHEL dataset demonstrate that our proposal achieves a \textit{recall@64} of 84.72\%, a 2.66 points improvement compared to previous best results.
\end{abstract}





\maketitle

\section{Introduction}

\begin{figure}[!t]
	\centering
	\includegraphics[width=0.95\linewidth]{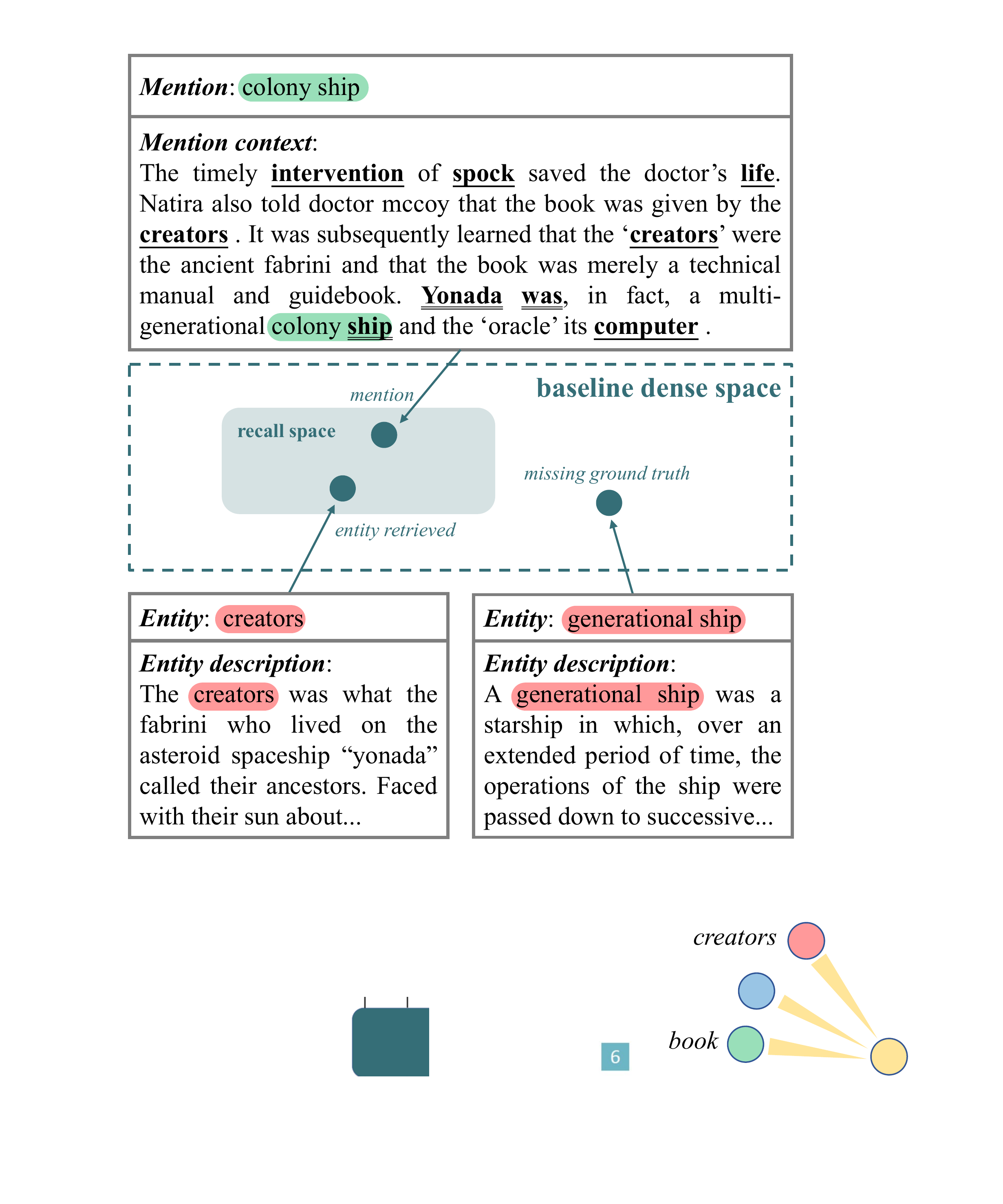}
	\caption{
% 	The embedding of mention \textit{colony ship}, retrieved entity \textit{creators} and missing ground truth entity \textit{generational ship} in dense space of baseline system BLINK.
    A bad case from baseline system BLINK.
    % Both mentions and entities are represented by the sentence embeddings from BERT.
    For mention \textit{clony ship}, the entity \textit{creators} is retrieved while the grouth truth \textit{generational ship} is missed.
	In the context of \textit{clony ship}, solid underline words~(e.g. \uline{creators}) note for the words with high attention in BLINK, and double underline~(e.g. \uuline{ship}) words for our GER framework.
	}
	\label{example}
	\vspace{-2em}
\end{figure}

Entity Linking~(EL) is a task of linking mentions in unstructured context to referent entries in a structured Knowledge Base~(KB) \citep{han2011collective}. 
% EL is an essential task in semantic text understanding and information extraction~(IE). 
% It can benefit other NLP tasks, such as question answering \citep{li-etal-2020-efficient}, text generation \citep{puduppully-etal-2019-data}, and dialogue system \citep{chen-etal-2017-robust}.
Previous systems on entity linking have achieved high performance given a large set of mentions and target entities for training. 
% Moreover, they typically use powerful resources such as a high-coverage alias table \citep{kolitsas-etal-2018-end}, and linking frequency statistics \citep{botha-etal-2020-entity}. 
However, such resources and labeled data may be limited and expensive in some domains, such as law and terminology. 
Without such massive training data, traditional entity-linking systems perform poorly.
% Therefore, domain-independent entity linking systems need to be developed. 
To evaluate the ability of the entity linking system to generalize to unseen entity sets, \citet{logeswaran-etal-2019-zero} proposed the zero-shot entity linking task and the ZESHEL dataset.
There are two assumptions under the \textbf{zero-shot setting}: (1) labeled mentions for the target domain are unavailable, (2) mentions and entities are only defined through textual descriptions~(a.k.a. \textbf{mention context} and \textbf{entity description}).
% In ZESHEL, each mention is extracted from a document in Wikia, and each entity corresponds to a document describing its details as shown in Figure \ref{example}.
%To build a baseline system, they adopt a two-stage pipeline including candidate entity retrieval using the BM25 algorithm and candidate entity ranking based on BERT \citep{devlin-etal-2019-bert}.
%previous work about the zero-shot entity linking task
Most zero-shot entity linking systems follow a two-stage pipeline: Candidate Entity Retrieval, where top-k candidate entities are retrieved based on scores such as inner product values of the mention vector and entity vector, and Candidate Entity Ranking, where the candidates are ranked to find the most probable one. 

In this paper, we focus on \textbf{first-stage entity retrieval}, since the overall accuracy is capped by recall performance in this stage.
Given the mention contexts and entity descriptions in the source domain simultaneously, a general way is to embed them in a dense space and calculate similarity scores to retrieve entities for the given mention.
\citet{wu-etal-2020-scalable}  propose a BERT-based \citep{devlin-etal-2019-bert} bi-encoder \citep{Humeau2020Poly-encoders:} model BLINK to encode the mention contexts and entity descriptions and utilize the sentence embedding to represent mentions and entities, followed by Maximum Inner Product Search~(MIPS) of the mention vector to find the closest $k$ candidate entities. 
The following embedding-based models \cite{yao-etal-2020-zero,partalidou2021improving} have widely adopted the BERT-based bi-encoder architecture, where the sentence embedding is defined as the output of the \texttt{[CLS]} token \footnote{In this paper, we refer to the sentence embedding as the output of \texttt{[CLS]} token following the recent models.}.

However, we argue that such coarse sentence embeddings can not fully model the information of mentions/entities.
% Specifically, such sentence embedding may be misled by other unrelated words with higher attention scores, leading to a shift in the semantic vector space. 
Intuitively, the sentence embeddings model the information of whole sentence rather than the mention/entity.
When the attention scores from the \texttt{[CLS]} token to mentions/entities are relatively low, such sentence embeddings may be misled by other high-attention words, leading to a shift in the semantic vector space. 
Figure \ref{example} shows a bad case from BLINK.
Both mentions and entities are represented by sentence embeddings.
For mention \textbf{colony ship}, the attention scores from the \texttt{[CLS]} token to \textit{colony ship} are relatively low, while \textit{creators} receives higher attention.
Hence, the entity \textbf{creators} becomes the closest entity in the dense space, while ground truth entity \textbf{generational ship} is missed. 
% As shown in Table \ref{atten_distri},
% We further analyze the attention distributions in the test set of popular benchmark ZESHEL. 
% For 53.59\% samples, the attention scores from the \texttt{[CLS]} token to mentions are relatively low.
% For each sample, we rank the input tokens by their attention scores from the \texttt{[CLS]} token and record the ranking of mention.
% As shown in Table \ref{atten_distri}, for 53.59\% samples, the attention scores towards the mention rank in [96,128) among the 128 input tokens.
%locates at the outside of the recall space.
% To understand the mention \textbf{colony ship}, in addition to the sentence embedding, the model should capture extra fine-grained information from the mention words themselves and other related words such as \texttt{generational} and \texttt{ship}.
% Therefore, we should model more fine-grained information to get better representation. 
% Meanwhile, since the sentence embeddings can not fully model the information of mentions/entities, 
Moreover, one intuitive idea is to adopt the output of mentions/entities from BERT rather than the \texttt{[CLS]} token.
However, the output of mentions/entities is highly similar to the \texttt{[CLS]} token due to the over-smoothing problem \citep{DBLP:journals/corr/abs-2202-08625}.
Therefore, more information about mentions/entities besides the output of BERT is required.
% Therefore, a better way to model the fine-grained information need to be developed.
%Moreover, the model should pay more attention to the related words of mention/entity such as \textit{ship} which are crucial to understanding mention. 
% Moreover, to capture more information about mentions/entities, the model should pay more attention to the semantically related words such as \textit{ship} and \textit{generational} in the context of \textbf{colony ship}.

%How people solve this problem in other domain

%Introduce the overall information about the method and show the main contribution
To address these issues, we propose a novel \textbf{G}raph enhanced \textbf{E}ntity \textbf{R}etrieval~(GER) framework.
Our key insight is to learn extra fine-grained information about mentions/entities as complementary to coarse-grained sentence embeddings. 
We extract the knowledge units as the information source and design a novel Graph Neural Network to aggregate these knowledge units.
Concretely, we employ the sequence prediction model \cite{Stanovsky2018SupervisedOI} to extract Subject-Predicate-Object~(SPO) triplets as knowledge units. 
% These knowledge units contain the core information of mention context/entity description.
After that, we build a graph by connecting the knowledge units to the corresponding mention/entity.
Such \textbf{graph design} allows mention/entity to \textbf{aggregate} information from these knowledge units.
To avoid the graph bottleneck\cite{DBLP:conf/iclr/0002Y21} for the central mention/entity node, we construct a \textbf{hierarchical} graph to reduce neighboring nodes and then design a novel Hierarchical Graph Attention Network (HGAN).
Finally, we employ the output of the central mention/entity node as fine-grained information, since they capture information at word level.

%The embedding of mention/entity node in the graph is utilized for word-level information.
%and sentence embedding of BERT are fused to get a better representation. 

% In experiments, we compare our graph enhanced encoder with the baseline combining sentence embedding and token embedding of the mention/entity word from BERT. 
% We find that both sentence-level embeddings and word-level embeddings are important for mention/entity representation, and the token embedding from BERT performs poorly at capturing the word-level information. 
% Empirical results on ZESHEL and other benchmarks show that our model outperforms the baseline.
We perform extensive experiments on Wikia based and Wikipedia based benchmarks, and compare GER with other strategies to learn fine-grained information.
Empirical results demonstrate that fine-grained information is beneficial to represent the mentions/entities, and the GER framework can capture such information better.
% Moreover, our GER framework achieves comparable results given 40\% training samples compare to the baseline system.
To evaluate overall performance, we also design the experiments for candidate ranking stage.
Compared to baseline BLINK, our GER framework achieves higher recall@64 with \textbf{slight extra time cost} in the zero-shot entity retrieval stage, but \textbf{saves more time} in the entity ranking stage.
% Further analysis implies that the HGAN can also improve the attention scores towards mention/entity and other related words.

%\item To the best of our knowledge, we are the first to adopt an entity-level graph for zero-shot entity retrieval, which can be easily transferred to benefit other tasks such as logic reasoning.
%\item To the best of our knowledge, we are the first to adopt an entity-level graph to model the word-level information for zero-shot entity retrieval, which can be easily transferred to benefit other tasks such as logic reasoning.
% \item We explore the mechanism to capture word-level embedding and further discover that both sentence-level information and word-level information play an important role in mention/entity representation.
% \item We propose GER framework, including a graph enhanced encoder to capture word-level information.
% \item We evaluate GER on the zero-shot entity linking dataset ZESHEL and Wikipedia-based entity linking datasets. The experimental results show that our model achieves significant improvements over the BERT-based baseline.
The main contributions of our paper are as follows:
\begin{itemize}
\item We show that coarse-grained sentence embeddings can not fully model the mentions/entities, and more fine-grained information is necessary for the zero-shot entity retrieval task.
% We find an issue that was overlooked for the zero-shot entity retrieval task, i.e., coarse-grained information such as sentence embedding can not fully model the mentions/entities due to attention bias, and more fine-grained information is necessary.

\item We propose GER, which learns extra fine-grained information about mentions/entities as complementary to coarse-grained sentence embeddings.
Particularly, we construct a mention/entity central graph and design a novel Hierarchical Graph Attention Network.

\item We evaluate GER on several zero-shot entity retrieval datasets, and experimental results demonstrate that our framework achieves significant improvements compared with previous models.

\end{itemize}

% \begin{figure*}[!t]
% 	\centering
% 	\includegraphics[width=0.9\linewidth]{figure/GEL_framework_v3.pdf}
% 	\caption{
% 	Overview of our GER framework.
% 	We use the paired $(M_{src}^i,E_{src}^i)$ from source domain to optimize mention encoder and entity encoder. 
% 	During inference phase, the embedding of all entities from target domain would be computed and cached.
% 	For each mention in target domain, we will retrieve the top-k nearest neighbor entities as candidates.
% 	}
% 	\label{framework}
% 	%\vspace{-2.0em}
% \end{figure*}

\section{Related Work}

\subsection{Zero-shot Entity Retrieval in Entity Linking}
% For the zero-shot entity linking, the entities in knowledge base can be as large as 5.9M which is time-consuming.
% Hence, most state-of-the-art models follow a two-stage framework: candidate retrieval stage and candidate ranking stage.
% They retrieve candidates in an extensive entity dictionary and rank the candidates retrieved to find the best one as a trade-off between accuracy and efficiency.
% For zero-shot entity linking, the overall accuracy is capped by recall performance in the candidate retrieval stage.
% On the other hand, for the candidate ranking stage, the cross-encoder \citep{Humeau2020Poly-encoders:} shows promising performance \citep{logeswaran-etal-2019-zero, wu-etal-2020-scalable,Tang_Sun_Jin_Zhang_2021}. 
% Therefore, recent works trend to improve the recall performance in the retrieval stage.

For candidate retrieval, two mainstream approaches are term-based, which focus on obtaining term-based representations, and semantically enriched, which utilize specific characteristics of entities~(attributes, types, and relationships) \citep{DBLP:series/irs/Balog18}.
For term-based, \citet{logeswaran-etal-2019-zero} use BM25 to measure the similarity between the mention word and entity description. 
% The coverage of the top-64 candidates is less than 77\%.
Other semantically enriched methods focus on learning better semantic representations. % for mentions and entities.
BLINK \citep{wu-etal-2020-scalable, yao-etal-2020-zero} employs a BERT-based bi-encoder \citep{Humeau2020Poly-encoders:} architecture to embed mentions/entities in semantic space via the sentence embeddings.
%However, most BERT-based models are based on a context window with 512 tokens while the length of entity description usually exceeds. 
% However, BERT allows 512 tokens at most while the length of entity description usually exceeds. 
% To capture long-range information, \citet{yao-etal-2020-zero} proposed an efficient position embedding initialization method to allow BERT to read 1024 tokens.
Another line is to inject more information for better representations.
\citet{partalidou2021improving} add an additional token before mention word, which is similar to BERT-based entity representations with entity type information \citep{poerner-etal-2020-e,jia-zhang-2020-multi} in other NLP applications. 
% To leverage rich semantics in medical knowledge graphs, \citet{kong2021zero} design graph-based tasks to incorporate the semantics of parent-child relationships into learning.
Moreover, mention-to-mention affinities can also provide more information to model mentions/entities \citep{DBLP:journals/corr/abs-2109-01242}.

%\citet{Tang_Sun_Jin_Zhang_2021} proposed a bidirectional multi-paragraph reading model to leverage more textual information and enhance text understanding capability.
%In this work, we try to learn an additional word-level representation from the same length of tokens as BLINK for efficiency.

Most existing work employs coarse-grained sentence embeddings to represent mentions/entities. In this work, we argue that such coarse-grained sentence embeddings can not fully model mentions/entities.
We design HGAN to capture the fine-grained embedding as complementary to get more comprehensive representations.
% we argue that such coarse-grained sentence embeddings can not fully model mentions/entities.

% \subsection{Graph Enhanced Text Representation}
% Using external knowledge graphs or syntax-based graphs to enhance the text representation has been extensively studied and applied in recent years.
% Specially, syntax-based graph extract from the input text can provide more syntax information \citep{li-etal-2021-dual-graph,tian-etal-2021-aspect,tang-etal-2020-dependency,huang-etal-2020-syntax} for aspect-based sentiment analysis task.
% In the text classification task, \citet{huang-etal-2019-text,zhang-etal-2020-every} build a graph for a given text by regarding its words as nodes and co-occurrences between words in sliding windows as edges. 
% After constructing the graph, they hire the GNNs such as GCN \citep{Kipf:2016tc} and GAT \citep{veli2018graph} to capture structural information at the word level.

% % Inspired by their work, we utilize the graph to get more fine-grained information. 
% However, such strategies aim to capture more information at the sentence level. 
% In our work, we construct a knowledge-aware graph by SPO triplets and aim to learn fine-grained information. 
% Moreover, compared to the GAT and GCN, our new designed HGAN can avoid the information bottleneck \cite{DBLP:conf/nips/WuRLL20,DBLP:conf/iclr/0002Y21} for central node through hierarchical architecture.

\subsection{Entity Embedding}
Entity embedding methods map the entity to vector representations from external knowledge sources such as structured Knowledge Graph~(KG) and unstructured corpus.
KG-based methods optimize the score of observed triplets.
These methods can be categorized as: distance models~\citep{DBLP:conf/nips/BordesUGWY13, DBLP:conf/iclr/SunDNT19, DBLP:conf/acl/ChamiWJSRR20}, which adopt distance-based scoring function, and linear models~\citep{DBLP:conf/nips/0007TYL19, DBLP:journals/jmlr/TrouillonDGWRB17}, which adopt similarity-based scoring function.
For unstructured corpus, the Pre-trained Language Models~(PLMs) \citep{DBLP:conf/naacl/DevlinCLT19, DBLP:journals/corr/abs-1907-11692} learn rich semantic information via pre-training tasks.
Moreover, KG-based model and PLM can be jointly optimized to enhance each other, including ERNIE \citep{DBLP:conf/acl/ZhangHLJSL19}, K-BERT \citep{DBLP:conf/aaai/LiuZ0WJD020}, KnowBERT \citep{DBLP:conf/emnlp/PetersNLSJSS19} and KEPLER \citep{DBLP:journals/tacl/WangGZZLLT21}.

Under the zero-shot entity retrieval setting, mentions and entities are only defined through textual descriptions (a.k.a. mention context and entity description).
Our GER aims to better represent mentions/entities adopting the textual descriptions without any external KG.

\begin{figure}[!t]
	\centering
	\includegraphics[width=0.95\linewidth]{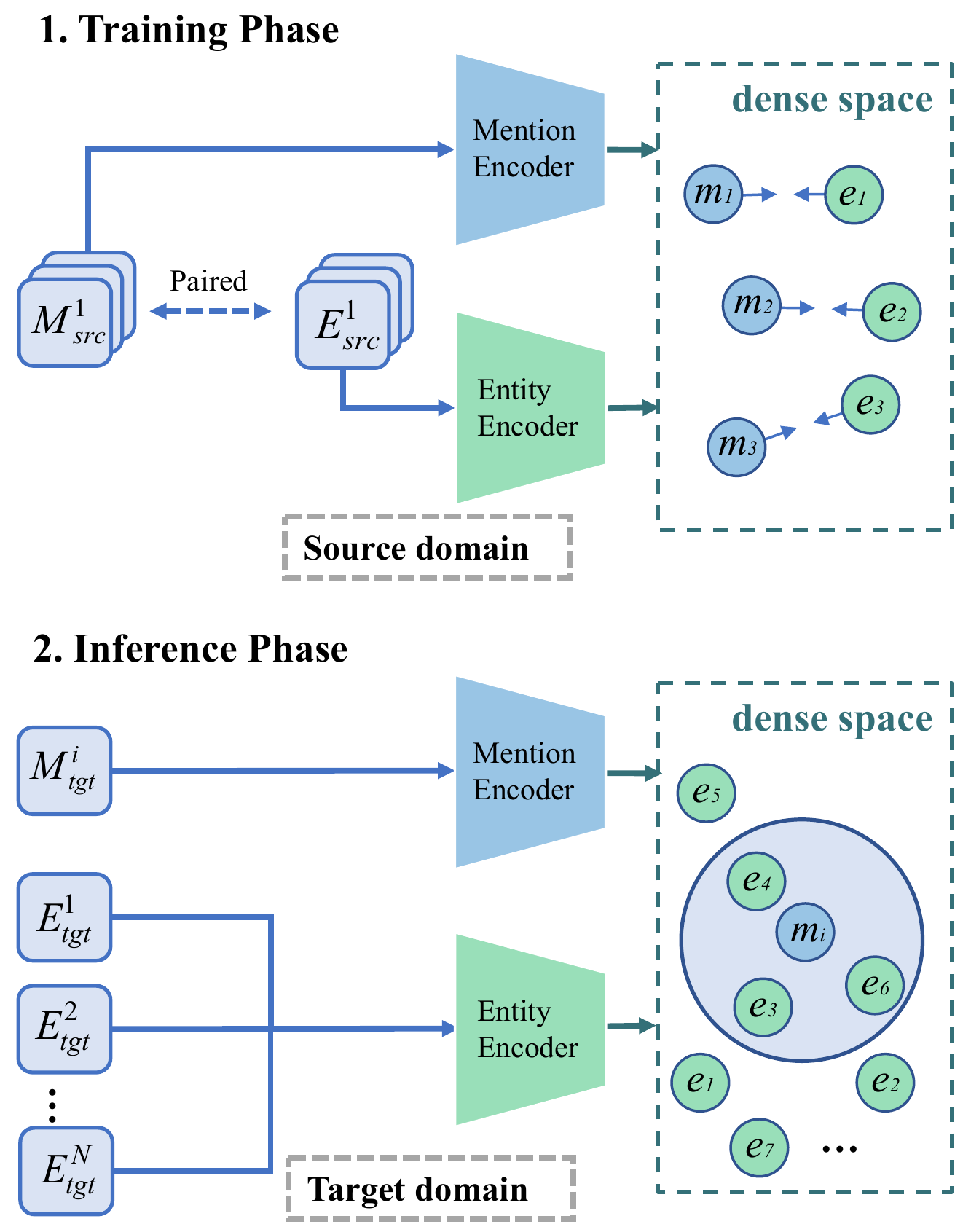}
	\caption{
	Overview of our GER framework.
% 	We use the paired $(M_{src}^i,E_{src}^i)$ from source domain to optimize mention encoder and entity encoder. 
% 	During inference phase, the embedding of all entities from target domain would be computed and cached.
% 	For each mention in target domain, we retrieve the top-k nearest neighbor entities as candidates.
	}
	\label{framework}
	\vspace{-1.0em}
\end{figure}

\section{Methodology}

\subsection{Task Formulation}
%Introduction the formulation of entity linking and zero-shot entity linking
\textbf{Entity Retrieval.}
% Given an unstructured text $c$ with a pre-identified mention $m$,
Given a mention $m$ and the corresponding mention context $c$, the goal of entity linking task is to find a paired entity $e$ from the knowledge base $E = \{e_1, e_2, \ldots, e_N\}$.
The size of the knowledge base $N$ can be very large, thus encoding those entities is time-consuming. % such as $5.9$M for Wikipedia, 
Hence, a common strategy is to consider entity linking in the two-stage paradigm: (i) retrieve $k$ candidate entities $\{e^1_i, e^2_i, \ldots, e^k_i\} $ for mention $m_i$, where $k \ll N$. (ii) rank these candidate entities. 
In this paper, we focus on candidate entity retrieval stage.

\textbf{Zero-shot Entity Retrieval.}
Under zero-shot setting, there is no graph information but corresponding entity description $d$ for each entity $e$. 
Moreover, we require $E_{src} \bigcap E_{tgt} = \emptyset$, where $E_{src}$ and $E_{tgt}$ mean the knowledge bases from source domain~(for train) and target domain~(for inference), respectively.

\subsection{Overall Framework}
Figure \ref{framework} shows the framework of GER.
Given the paired~(mention, entity) in the source domain for training, two individual encoders are jointly employed to map the mentions and entities into dense vectors, respectively. 
The training goal is to minimize the distance between the paired~(mention, entity).
% After that, the scores between each mention and each entity in one batch are calculated by their inner dot product.
% We optimize the symmetric entropy loss, which is similar to CLIP \citep{DBLP:conf/icml/RadfordKHRGASAM21}.
% For a batch of $N$ paired mention and entity, the goal is to maximize the scores of $N$ real pairs while minimizing the scores of $N^2-N$ incorrect pairs. 
For inference, we firstly employ the entity encoder to encode all entities from $E_{tgt}$ and cache these vectors, to achieve fast and real-time inference.
% two trained encoders are used to encode mentions separately and all entities in the knowledge base $E_{tgt}$ to dense vectors. 
For each mention, we compute the scores with all entities and retrieve top-$k$.

%Since the mention encoder and entity encoder share the same architecture, we introduce the details of the mention encoder as follows.

%This section introduces the details about our proposed GER framework and how we optimize it.
\begin{figure*}[!ht]
	\centering
	\includegraphics[width=0.95\linewidth]{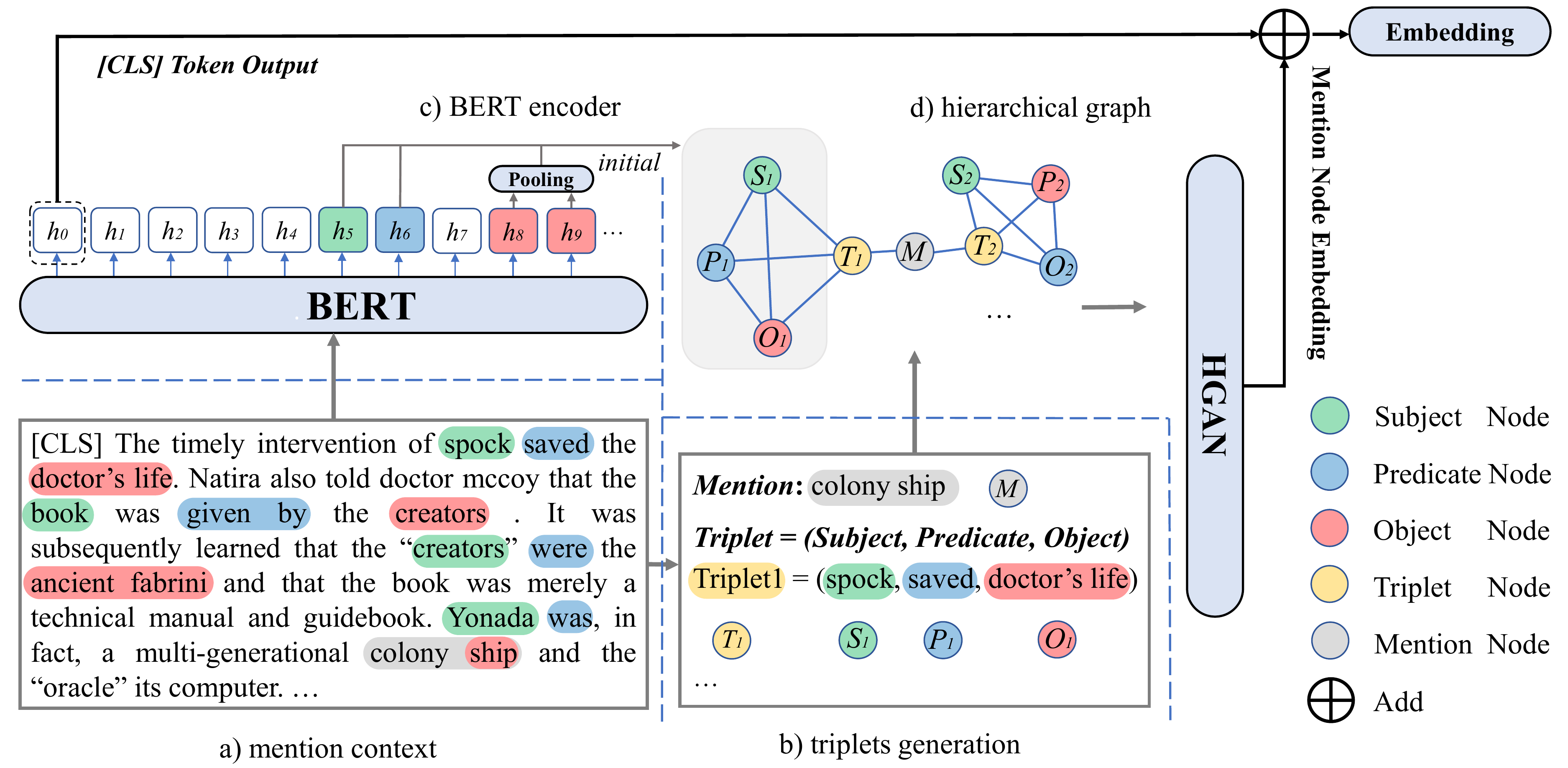}
	\caption{
%Overview of our graph enhanced encoder. For a given mention and its context (shown in part a), we extract the subject-predicate-object triples (shown in part b). BERT are used to encode the context (shown in part c)  and the triplets are used to construct a hierarchical graph (shown in part d). After initializing the node embedding by token embedding from BERT, we adopt a novel Hierarchical Graph Attention Network (HGNN) to learning the graph embedding. 
Overview of mention encoder in GER. 
% Mention encoder and entity encoder follows the same architecture.
For the given mention representation~(shown in part a), we extract the triplets~(shown in part b, green for the subject, blue for the predicate, and red for the object) as knowledge units. 
To avoid the graph bottleneck \cite{DBLP:conf/iclr/0002Y21}, we add a triplet node~(in yellow) between the mention/entity node and each triplet, and thus build the hierarchical graph~(shown in part d).
% We design the HGAN to encode the hierarchical graph and utilize the embedding of mention node~(entity node for entity encoder) as fine-grained information. 
}
	\label{encoder}
	\vspace{-1em}
\end{figure*}

\subsection{Mention/Entity Encoder}
Following the bi-encoder setting, the mention encoder and entity encoder share the same structure.
Figure \ref{encoder} shows the overview of mention encoder.
The goal is to learn extra fine-grained information about mentions as complementary to sentence embeddings.
% We extract the knowledge-aware units from mention context or entity description to construct a graph. 
% For graph nodes, the node embeddings are initialized by the corresponding token embeddings from PLM such as BERT. 
% After applying the HGAN, we employ the node embedding of the mention/entity node as fine-grained information and fuse it with sentence embedding from BERT to get a more comprehensive representation.

\textbf{Hierarchical Graph.}
% To better understand the mention/entity at the word level, we adopt a simple yet effective way to extract the Subject-Predicate-Object~(SPO) triplets from the mention context/entity description. 
% For the given context, we use the parsing tool Spacy\footnote{https://spacy.io/} to get the syntax dependency tree and part-of-speech tag for each word.
% After that, we design special rules to recognize the SPO triplets based on grammar prior knowledge. 
% The details are explained in Appendix \ref{appendix_spo_pipeline}.
The goal is to extract the core information~(a.k.a, knowledge units) of mention context~(entity description) and aggregate these information to mention~(entity).
We extract the Subject-Predicate-Object~(SPO) triplets as knowledge units following a two-stage strategy: 1) first, we employ the open information tool trained on semantic role labeling task from allennlp \footnote{https://demo.allennlp.org/open-information-extraction} to recognize the semantic role for each word.
2) later, we design special rules to recognize the SPO triplets based on grammar prior knowledge. 
After that, we construct a mention/entity centralized graph by connecting these SPO triplets to central mention/entity node.
% Moreover, the mention/entity will be treated as a special central node to gather knowledge.
% For each knowledge-aware unit, a virtual node is added as a relay node between SPO nodes and central mention/entity node.
To avoid the graph bottleneck \cite{DBLP:conf/iclr/0002Y21}, we add a triplet node between the mention/entity node and each triplet. 
Hence, we can reduce the neighboring nodes for central mention/entity node from $3N$ to $N$, where $N$ is the number of triplets.

\textbf{Text Modeling.}
% After constructing the hierarchical graph, we initialize the nodes with the output of the corresponding BERT tokens.
% As shown in Figure \ref{encoder}, for the triplet \textit{(spock, saved, doctor's life)}, the node \textit{spock} is initialized with the token output of spock from BERT.
% Meanwhile, we adopt the mean pooling operation for nodes that correspond to more than one token, such as the node \textit{doctor's life}.
%The context of mentions and entities provide the information which can help us understand their meaning. 
We encode the mention~(given mention $m$ and its context $c$) and entity~(given entity $e$ and its description $d$) as follows:
%both the representation of mention and the representation of entity:
\begin{equation}
    Y_m = T_m(\texttt{[CLS]} \ c_{l} \  \texttt{[MS]} \ m \  \texttt{[ME]} \ c_{r}) 
\end{equation}
\begin{equation}
    Y_e = T_e(\texttt{[CLS]} \ e \ \texttt{[ENT]} \ d )
\end{equation}
where $m$, $c_l$, $c_r$, $e$, $d$ are the word-pieces tokens of mention, context before mention, context after mention, entity title and corresponding description, respectively.
\texttt{[MS]},\texttt{[ME]} are the special tokens to mark the start and end of mention span.
\texttt{[ENT]} serves as the delimiter of entity and its descriptions.
$T_m$ and $T_e$ are two independent encoders.
The baseline BLINK adopts the sentence embedding~(output of the \texttt{CLS} token, such as $Y_m[0]$ and $Y_e[0]$) to represent mentions/entities.

% where $\tau_m$ and $\tau_e$ means the input of mention and entity respectively, $T_m$ and $T_e$ are two independent BERT. 
% For mention word $m$ and mention context $c$, the representation is composed of the word-pieces surrounding the mention and mention word itself:
% % Specifically, we construct $\tau_m$ for each mention example as:
% \begin{equation}
%     \tau_m = \texttt{[CLS]} \ c_{l} \  \texttt{[MS]} \ m \  \texttt{[ME]} \ c_{r}
% \end{equation}
% where $m$, $c_l$, $c_r$ are the word-pieces tokens of mention, context before mention and context after mention, respectively, and \texttt{[MS]},\texttt{[ME]} are the special tokens to mark the start and end of mention span.
% %For entity, $\tau_e$ are initialized by the concatenation of entity title and corresponding description, which can be formulated as:
% For entity $e$ and its description $desc$, $\tau_e$ can be formulated as:
% \begin{equation}
%     \tau_e = \texttt{[CLS]} \ t \ \texttt{[ENT]} \ desc 
% \end{equation}
% where $t$, $desc$ are the word-pieces tokens of entity title and corresponding description, \texttt{[ENT]} serves as the delimiter of entity titles and descriptions.
% Hence, we get $Y_m$, the embeddings for mention $m$, and $Y_e$, the embeddings for entity $e$. 

\textbf{Hierarchical Graph Attention Network.}
We design the Hierarchical Graph Attention Network~(HGAN) to model the fine-grained information for mentions/entities.

As shown in Figure \ref{encoder}, for each triplet, the subject, predicate, and object are all viewed as individual nodes and connected to each other.
% Besides, a virtual relay node is added for each triplet, which serves as the information bridge between SPO nodes and the mention/entity node. 
To avoid the graph bottleneck, we add the triplet node between the SPO nodes and the central mention/entity node.
For SPO nodes and mention/entity nodes, we initialize their representation by aligning to the output of corresponding tokens. 
Taking the mention nodes for example, we initialize the node representation by:
\begin{equation}
    \mathbf{h}^{0}_{m} = red(Y_m[p_{start}:p_{end}])
\end{equation}
where $p_{start}$, $p_{end}$ are the start index and the end index of the mention token among all input tokens, and $red(\cdot)$ is a readout function reducing the sequence of vectors into one.
We choose $red(\cdot)$ to be average pooling.
%the common pooling methods including average pooling and max pooling.

However, the triplet nodes do not correspond to tokens among the input tokens.
In GER, we initialize the representation of the triplet nodes by gathering the information of corresponding SPO nodes.
% gather the information of corresponding SPO nodes to initialize the 
% To initialize the representation of the virtual triplet node, we gather the information of corresponding SPO nodes.
Let $\mathbf{h}^{0}_{s} \in \mathbb{R} ^{1 \times d}$, $\mathbf{h}^{0}_{p} \in \mathbb{R} ^{1 \times d}$ and $\mathbf{h}^{0}_{o} \in \mathbb{R} ^{1 \times d}$ note for the node representation of subject, predicate, and object respectively, the representation of the corresponding triplet node can be calculated by:
\begin{equation}
    \mathbf{h}^{0}_{t} = \left[ \mathbf{h}^{0}_s \| \mathbf{h}^{0}_p \| \mathbf{h}^{0}_o \right] {\bf W}^{triple}
\end{equation}
where $\|$ notes the concatenation operation and $\mathbf{W}^{triple} \in \mathbb{R}^{3d\times d}$ is a learnable matrix.

%Considering the triplet as the potential information to get the point of mention (or entity), we need to fuse the information of the nodes, which helps us understand the mention (or entity). 
After constructing the graph and acquiring corresponding initial representation for each node, we follow the idea of GAT \citep{veli2018graph} and apply the multi-head attention mechanism to aggregate the information from neighborhood nodes.
Specifically, in an $L$ layer HGAN, for each layer, we update the representation $\mathbf{h}^{(l)}_i \in \mathbb{R}^{1\times d}$ of each node $i$ by:
\begin{equation}
    \mathbf{h}^{(l)}_i = \sigma \left( 
    \frac{1}{K}
    \sum_{k=1}^{K}
    \sum_{ j \in \mathcal{N}_i \cup \{i\}} \alpha_{ij}^{k}  \mathbf{h}^{(l-1)}_j {\bf W}^k \right)
\end{equation}
where $\mathcal{N}_i$ are the neighbor nodes of node $i$, $K$ is the number of attention heads, ${\bf W}^k \in \mathbb{R}^{d \times d}$ are weight matrix for each head, $\alpha_{ij}^k$ means the attention score between node $i$ and node $j$, and $\sigma(\cdot)$ is an activate function such as $relu(\cdot)$. 
Attention scores $\alpha_{i,j}^{k}$ for head $k$ are calculated by:
\begin{equation}
\label{atten_equation}
\alpha_{ij}^{k} = \frac{\exp(
e_{ij}^{k}
)}{\sum_{j^{'} \in \mathcal{N}_i \cup \{i\}} \exp(
e_{ij^{'}}^{k}
)}
\end{equation}
\begin{equation}
e_{ij^{'}}^{k} = \texttt{LeakyReLU} ([\mathbf{h}_i {\bf W}^{k} \| \mathbf{h}_{j^{'}} {\bf W}^{k} ] {\bf a} )
\end{equation}
where $\texttt{LeakyReLU}(\cdot)$ is an activation function and ${\bf a} \in \mathbb{R} ^{2d \times 1}$ is a score matrix.

\textbf{Feature Fusion.} 
% Combining graph level information and textual information can help us model mention(or entity) better. 
% For graph level information $\mathbf{V}_{graph}$, we use the node representation of mention (or entity) $\mathbf{H}^{L}_m$ at last layer. 
% The representation of $\texttt{[CLS]}$ are employed as textual information $\mathbf{V}_{text}$. 
% Hence, we get the final embedding of our graph enhanced encoder:
% \begin{equation}
%     \mathbf{V}_{final} = \mathbf{V}_{text} + \mu \mathbf{V}_{graph}
% \end{equation}
% where $\mu$ is a hyper parameter and $\mathbf{V}_{final}\in \mathbf{R}^{1\times d}$.
%To fuse the sentence level embedding and the word level embedding, we follow the strategy to add them.
After capturing the fine-grained information via HGAN, we fuse the fine-grained embedding with the sentence embedding to obtain a more comprehensive representation.
% After applying the HGAN, we employ the output of mention/entity node as fine-grained information. 

For coarse-grained sentence embedding $v^{sen} \in \mathbb{R}^{1\times d}$, following the baseline BLINK, we use the output of the \texttt{[CLS]} token which is the first token in the input sequence of BERT. Take the mention $m$ for example:
\begin{equation}
    v^{sen} = Y_m[0]
\end{equation}
where $Y_m \in R^{L \times d}$ and $L$ is the length of the input sequence for $m$.

For fine-grained word-level embedding $v^{graph} \in \mathbb{R}^{1\times d}$, we utilize the node output of the mention node $m$ after applying the $L$ layer HGAN:
\begin{equation}
    v^{graph} = h^{L}_{m}
\end{equation}
For entity encoder, we apply a similar strategy utilizing the node output of the entity node as fine-grained embedding.

% Hence, we can get the fused embedding for mention/entity by:
In GER, we fuse the coarse-grained sentence embedding $v^{sen}$ and fine-grained embedding $v^{graph}$ by a gate mechanism:
\begin{equation}
\label{fuse}
    v = v^{sen} + \lambda v^{graph}
\end{equation}
where $\lambda$ is a learnable hyperparameter and $v \in \mathbb{R}^{1\times d}$.
In this way, the baseline BLINK is \textbf{a special case} of GER when we fix $\lambda$ as 0.

\subsection{Optimization and Inference}
% Under the zero-shot setting, we optimize the GER framework by the paired (mention, entity) from the source domain and evaluate the retrieval performance on the target domain.
For training, we optimize the GER framework by the paired (mention, entity) from the source domain.
Considering the time efficiency and training speed, we adopt the batched negative sampling strategy \citep{DBLP:conf/mlsys/LererWSLWBP19, DBLP:conf/aaai/WuBY22, DBLP:conf/naacl/Ding0FZJYTTLGBM22} where the other samples at the same batch are viewed as negative samples to calculate the loss. 
Concretely, for a batch of randomly sampled mention-entity pairs $\left( 
(m_1,e_1),...(m_{bsz},e_{bsz}) \right)$ where $bsz$ notes for the batch size, the loss is computed as:
\begin{equation}
    \mathcal{L}(m_i,e_i) = \mathcal{L}_1(m_i,e_i) + \mathcal{L}_2(m_i,e_i)
\end{equation}
\begin{equation}
  \mathcal{L}_1(m_i,e_i) = -s(m_i,e_i) + \log \sum^{bsz}_{j=1} \exp(s(m_i,e_j)) 
\end{equation}
\begin{equation}
  \mathcal{L}_2(m_i,e_i) = -s(m_i,e_i) + \log \sum^{bsz}_{i=1} \exp(s(m_i,e_j))
\end{equation}
\begin{equation}
\label{scores}
    s(m_i,e_i) = v_{m_i} \cdot v^T_{e_i}
\end{equation}
% \begin{equation}
% \begin{split}
%     \mathcal{L}(m_i,e_i) &= -s(m_i,e_i) + \log \sum^{bsz}_{j=1} \exp(s(m_i,e_j)) \\ + \log \sum^{bsz}_{j=1} \exp(s(m_j,e_i))
% \end{split}
% \end{equation}
where $v_{m_i}$ and $v_{e_i}$ are the fused embeddings of mention $m_i$ and entity $e_i$ from the mention encoder and entity encoder, respectively.  

% \subsection{Inference}
At inference time, we will pre-compute and cache the entity representation for all the candidates, to achieve fast and real-time inference. 
As shown in Figure \ref{framework}, we retrieve $k$ candidate entities for each mention based on the scores defined by Equation \ref{scores}.

\section{Experiments}
In this section, we perform an empirical study of GER framework on several challenging datasets for zero-shot entity retrieval. 
We further analyse the GER performance for candidate ranking and do ablation studies about the fine-grained information.

\begin{table}[!h]
\resizebox{\linewidth}{!}
{
\begin{tabular}{ccccc}
\toprule
\textbf{KB} &
  \textbf{Dataset} &
  \textbf{Usage} &
  \textbf{\begin{tabular}[c]{@{}c@{}}Samples\\ Num\end{tabular}} &
  \textbf{\begin{tabular}[c]{@{}c@{}}Entity\\ Num\end{tabular}} \\ \midrule
\multirow{4}{*}{\begin{tabular}[c]{@{}c@{}}Wiki-\\ pedia\end{tabular}} &
  \multirow{2}{*}{AIDA} &
  Train &
  18,317 &
  \multirow{4}{*}{5,903,530} \\
                       &                         & Valid & 4,763  &         \\ \cline{2-4}
                       & WNED-CWEB               & Test  & 10,392 &         \\ \cline{2-4}
                       & AQUAINT               & Test  & 678  &         \\ \midrule
\multirow{3}{*}{Wikia} & \multirow{3}{*}{ZESHEL} & Train & 49,275 & 332,632 \\
                       &                         & Valid & 10,000 & 89,549  \\
                       &                         & Test  & 10,000 & 70,140  \\ \bottomrule
\end{tabular}
}
\caption{
Statistics of entity retrieval datasets and knowledge base, samples num means the size of paired mentions and entities.
For each KB, we use the corresponding train dataset~(e.g., AIDA train set) to optimize our GER framework, and report the recall results on test dataset~(e.g., WNED-CWEB).
}
\label{dataset}  
\vspace{-2.0em}
\end{table}

\begin{table*}[ht]
%\resizebox{\linewidth}{!}
%{
\centering
\begin{tabular}{l|ccccccc}
\toprule
\textbf{Method} & \textbf{R@1}   & \textbf{R@4}   & \textbf{R@8}     & \textbf{R@16}  & \textbf{R@32} & \textbf{R@50}  & \textbf{R@64}  \\ \midrule
BM25 \cite{logeswaran-etal-2019-zero}$^{\dagger}$                      & -     & -     & -  & -     & -  & -    & 69.13 \\
BLINK \cite{wu-etal-2020-scalable}$^{\dagger}$  & -     & -  & -   & -     & -       & -     & 82.06 \\
\citet{partalidou2021improving}$^{\dagger}$  & -     & -     & -     & -   & -  & 84.28     & - \\
ARBORESCENCE \citep{DBLP:journals/corr/abs-2109-01242}$^{\dagger}$  & -     & -     & -     & -    & -     & -     & 85.11 \\
BLINK \cite{wu-etal-2020-scalable}$^{*}$                    & 38.01 & 62.08 & 69.19 & 75.39 & 80.03 & 82.69 & 83.98 \\ \midrule
BERT Mean Pooling         & 33.65 & 57.74 & 65.17 & 71.38 & 75.85 & 78.66 & 80.14 \\
BERT Max Pooling          & 36.94 & 60.42 & 68.34 & 73.83 & 78.40 & 81.09 & 82.65 \\ \midrule
BLINK + BERT Mean Pooling & 34.12 & 58.41 & 66.19 & 72.24 & 76.93 & 79.79 & 81.16 \\
BLINK + BERT Max Pooling  & 38.45 & 63.46 & 70.68 & 76.72 & 81.11 & 83.63 & 84.83 \\ \midrule
GER~(ours)       & \textbf{42.86} & \textbf{66.48} & \textbf{73.00} & \textbf{78.11} & \textbf{82.15} & \textbf{84.41} & \textbf{85.65} \\ \bottomrule
\end{tabular}
%}
\caption{\textit{Recall@K}~(R@K) results on the test set of ZESHEL dataset, which is the average of 5 runs with different random seeds.
%M/E means the mention and entity. 
$^*$ notes for the results we reproduce.
$^{\dagger}$ notes for the results taken from their papers.
Best results are shown in bold.
GER outperforms all baselines significantly with paired t-test at p < 0.05 level considering R@64.
% For each dataset, \textbf{bold} indicate the best model.
%while \underline{underline} indicates the second best. 
}
\label{zeshel_res}  
\vspace{-1em}
\end{table*}

\begin{table*}[ht]
% \resizebox{\linewidth}{!}
% {
\centering
\begin{tabular}{l|ccc|ccc}
\toprule
\multirow{2}{*}{\textbf{Method}} & \multicolumn{3}{c|}{\textbf{WNED-CWEB}}          & \multicolumn{3}{c}{\textbf{AQUAINT}}             \\
                                 & \textbf{R@10}  & \textbf{R@30}  & \textbf{R@128} & \textbf{R@10}  & \textbf{R@30}  & \textbf{R@128} \\ \midrule
BLINK$^{*}$                     & 80.16 & 84.48 & 89.22 & 93.95 & 96.76 & 98.23 \\ \midrule
BERT Mean Pooling         & 79.87 & 84.79 & 89.35 & 94.54 & 96.90 & 98.23 \\
BERT Max Pooling          & 77.62 & 83.56 & 88.57 & 93.07 & 95.87 & 97.94 \\ \midrule
BLINK + BERT Mean Pooling & 80.13 & 84.33 & 88.81 & 94.84 & 96.31 & 98.23 \\
BLINK + BERT Max Pooling  & 78.75 & 84.16 & 88.81 & 93.22 & 96.02 & 97.35 \\ \midrule
GER (ours)                        & \textbf{80.79} & \textbf{85.34} & \textbf{90.13} & \textbf{95.28} & \textbf{97.05} & \textbf{98.82} \\ \bottomrule
\end{tabular}
% }
\caption{
\textit{Recall@K} (R@K) on dataset WNED-CWEB and AQUAINT. 
% Best results are shown in bold.
% $^*$ notes for the results we reproduce.
The experiments are all under the zero-shot settings. that entities are only defined by textual description and the entities in test set are unseen during training.
}
\label{benchmark_res}  
\vspace{-1.0em}
\end{table*}

\subsection{Datasets}

We evaluate our framework GER under two popular public knowledge bases: Wikia \footnote{https://www.wikia.com} and Wikipedia \footnote{https://www.wikipedia.org/}.
Statistics of datasets are listed in Table \ref{dataset}.

\textbf{Wikia Based.}
Wikias are community-written encyclopedias, each specialized in a particular subject such as film series and computer game. 
ZESHEL dataset \citep{logeswaran-etal-2019-zero} was constructed based on the documents in Wikia from 16 domains.
% which is the prevailing benchmark for zero-shot entity retrieval, was constructed based on the documents in Wikia from 16 domains.
For each domain, there are paired (mention, entity) and individual entity dictionary.
Following the baseline BLINK, we divided the 16 domains into: 8 domain for train, 4 for valid, and 4 for test.
As shown in Table \ref{dataset}, there are 49K, 10K, 10K paired examples in the train, valid, test sets, respectively. 
The candidate entity dictionary for each domain ranges in size from 10K to 100K.

\textbf{Wikipedia Based.}
We use the May 2019 English Wikipedia dump from KILT \citep{petroni-etal-2021-kilt} as our knowledge base, which includes 5.9M entities from Wikipedia. 
To the best of our knowledge, there is no standard zero-shot retrieval linking dataset in the full Wikipedia setting.
Therefore, in order to evaluate the performance under Wikipedia KB, we use standard entity linking datasets from different domains to satisfy the zero-shot setting.
Concretely, we employ the AIDA CoNLL-YAGO \citep{DBLP:conf/emnlp/HoffartYBFPSTTW11} for training, WNED-CWEB~\citep{DBLP:journals/semweb/GuoB18} and AQUAINT for test.
We preprocess the datasets to guarantee that the entities in test set are unseen in the training phase.

\subsection{Implementation}

We conduct experiments on both BERT-base and BERT-large following the baseline model for fair comparison. 
The max length for mention context and entity description are both set to 128.
For HGAN, we stack 3 layers with 8 attention heads. 
The learnable hyperparameter $\lambda$ is initial as $0.5$. 
All parameter are optimized by AdamW \citep{DBLP:conf/iclr/LoshchilovH19}, with learning rate 2e-5, gradient clipping 1.0 \citep{DBLP:conf/icml/PascanuMB13}, and 10\% warmup steps.
For ZESHEL, the batch size for training is 128 following the BLINK and training 5 epoches based on BERT-base takes 0.8h on 4 Tesla-A100 GPUs. 
For AIDA CoNLL-YAGO, we use the BERT-large model provided in BLINK as backbone, which is pretrained on 9M paired Wikipedia samples.
We set the batch size as 64 and train 10 epoches, which takes 1.4h on 4 Tesla-A100 GPUs.

\subsection{Comparison Methods}
For a fair comparison, we choose BLINK ~\citep{wu-etal-2020-scalable} as our baseline.
% , which limits the mention context length and entity description length to 128 as same as GER. 
Moreover, we also compare our GER framework with two strategies to learn fine-grained embedding~(BERT Mean/Max Pooling).
%including mean pooling and max pooling of mention/entity token embeddings. 

\textbf{BM25 \citep{logeswaran-etal-2019-zero} .} A frequency-based method to retrieve the candidates.

\textbf{BLINK \citep{wu-etal-2020-scalable} .} A BERT-based bi-encoder model where the mentions and entities are represented by the sentence embedding from BERT of the corresponding context.

\textbf{\citet{partalidou2021improving}.} They extend the BLINK by introducing a new pooling function and incorporating entity type side-information.

\textbf{ARBORESCENCE \citep{DBLP:journals/corr/abs-2109-01242}.} A model that fully utilizes mention-to-mention affinities by building minimum arborescences to make linking decisions.

\textbf{BERT Mean Pooling. } Compare to BLINK, we utilize the mean pooling of mention/entity token outputs as representation. 

\textbf{BERT Max Pooling. } Compare to BLINK, we utilize the max pooling of mention/entity token outputs as representation. 

\textbf{BLINK + BERT Mean Pooling. } We utilize the mean pooling of mention/entity token outputs as fine-grained information and fuse it with sentence embedding.

\textbf{BLINK + BERT Max Pooling. } We utilize the max pooling of mention/entity token outputs as fine-grained information and fuse it with sentence embedding.

\subsection{Main Results}

We conducted the experiments on Wikia-based and Wikipedia-based datasets. 
Table \ref{zeshel_res} and Table \ref{benchmark_res} report the recall results of the baseline models and GER.
Some findings are summarized as follows:

(1) Our proposed GER outperforms all baselines on the evaluation metrics for Wikia based datasets and Wikipedia-based datasets, which demonstrates the effectiveness of GER on getting more comprehensive representations for mentions/entities.
Compared to baseline BLINK, GER significantly improves the recall score with the paired t-test at p < 0.05 level considering recall@64.
In particular, GER achieves 85.65 recall@64 score in ZESHEL, which is 1.67 higher than BLINK we reproduce. 
% Moreover, our GER also performs better than ARBORESCENCE \cite{DBLP:journals/corr/abs-2109-01242} which models the mention-to-mention relation.
Following BLINK, for Wikipedia based datasets, we pretrain the BERT-large on 9M paired Wikipedia samples.
Hence, the results for baseline and GER are \textbf{high}.
However, GER still achieves higher results, such as a recall@10 of 95.28 for AQUAINT.

(2) The mean pooling and max pooling of mention/entity tokens perform worse than sentence embedding.
% We consider the mean pooling or max pooling of mention/entity tokens as \textbf{fine-grained embedding from BERT}.
Intuitively, BERT model is pre-trained on MLM and NSP tasks, where the output of [CLS] is employed to represent the whole sentence.
Thus, such fine-grained word-level embedding from BERT can not model the sentence due to the gap between the pretrain tasks and zero-shot entity retrieval task.
As shown in Table \ref{zeshel_res} and \ref{benchmark_res}, BERT Mean Pooling and BERT Max Pooling perform worse than BLINK, such as 33.65 and 36.94 compared to 38.01 for recall@1 on ZESHEL.

(3) Both fine-grained information and coarse-grained sentence embedding are crucial to represent mentions/entities.
For BERT Mean Pooling strategy, the recall@64 score on ZESHEL improves from the 80.14 to 81.16 if we fuse it with the sentence embedding.
On the other hand, the BLINK + BERT Max Pooling strategy achieves 0.85 higher than BLINK on recall@64 for the ZESHEL dataset.
For AQUAINT, as shown in Table \ref{benchmark_res}, we can also get the same conclusion such as 94.84 recall@10 for BLINK+BERT Mean Pooling compared to 93.95 for BLINK.

\label{mark}
(4) Our proposed HGAN can capture better fine-grained information than BERT.
As mentioned in (2), we consider the BERT Max/Mean Pooling as the fine-grained information from BERT.
In GER, we extract the knowledge units and capture the fine-grained information by HGAN.
As shown in Table \ref{zeshel_res} and \ref{benchmark_res}, our GER performs better than BLINK + BERT Mean Pooling and BLINK + BERT Max Pooling from recall@1 to recall@100 on all datesets.

\subsection{Ablation Studies}

To better understand our proposed GER framework, we conduct ablation experiments and compare our HGAN with other strategies to learn fine-grained information.
% utilizing sentence-level information and word-level information separately. 
% Moreover, after getting the SPO triplets, we try two strategies to encode these nodes: 1) apply to mean pooling on the node embeddings 2) connect the SPO nodes to mention/entity nodes and adopt a GAT network with the same layers and attention heads.
% Table \ref{ablation_study} reports the results on ZESHEL. We can observe that both sentence-level information and word-level information are important for getting a better representation. Comparing the results of sentence-level only and word-level only, we find that the sentence-level information is more crucial for our task.
% Moreover, compared to the baseline, we can observe that adopting the GAT can get a better result, while simply adding the node embedding brings a slight drop.
% It indicates that the triplets we extract contain the word-level information to get the point of mention/entity, and encoding the structural information also matters.
We guide our experiments with following research questions~(RQs):
\begin{itemize}
\item[\textbf{RQ1:}]
Can we use fine-grained information only~(without the sentence embedding) to represent mentions/entities ?
\end{itemize}

\begin{table}[h]
\resizebox{\linewidth}{!}
{
\begin{tabular}{l|l|cccc}
\toprule
Sentence-level & Word-level & R@1   & R@8 & R@32  & R@64  \\ \midrule
BERT           & -          & 38.01 & 69.19 & 80.03 & 83.98 \\
-              & HGAN       & 37.37 & 63.77 & 73.19 & 77.29 \\ \midrule
BERT           & Node Mean  & 37.29 & 69.62 & 80.15 & 83.88 \\
BERT           & GAT        & 39.23 & 70.07 & 80.14 & 84.09 \\ \midrule
BERT           & HGAN       & 42.86 & 73.00 & 82.15 & 85.65 \\ \bottomrule
\end{tabular}
}
\caption{The ablation study results of our GER~(BERT+HGAN) on ZESHEL dataset.
% For our proposed GER, we combine the sentence-level information from BERT and word-level information from HGAN.
}
\label{ablation_study}
\vspace{-2.0em}
\end{table}

We compare the results of GER with BERT only and HGAN only based on the ZESHEL dataset.
As shown in Table \ref{ablation_study}, we can observe that utilizing the fine-grained information only~(HGAN) will lead to a drop, especially for recall@64.
We consider this drop due to \textbf{gap in tasks} since BERT is pre-trained by MLM and NSP tasks rather than graph tasks.
Actually, sentence embedding from BERT is also vital for mention/entity representation.

\begin{itemize}
\item[\textbf{RQ2:}]
Is HGAN more suitable for modeling fine-grained information from knowledge-aware graph ? Why not use GAT ?
\end{itemize}
After constructing the knowledge-aware hierarchical graph, we try two strategies to model fine-grained information: 

\textbf{Node Mean.} Regardless of the structural information in the graph, we adopt the mean pooling of all nodes including the SPO nodes and the mention/entity node. 

\textbf{GAT.} We connect the SPO nodes to mention/entity node directly and apply the GAT \cite{veli2018graph} to model the fine-grained information.

As shown in Table \ref{ablation_study}, we can observe that applying the GAT performs better than the baseline, ranging from recall @ 1 to recall @ 64, while applying the Node Mean brings a slight drop for recall@64.
We can conclude that structural information in knowledge units is vital to modeling fine-grained information.
Moreover, HGAN performs better than GAT, indicating the \textbf{necessity for hierarchical design}.
GNNs such as GAT surfer from the \textbf{graph bottleneck} \cite{DBLP:conf/iclr/0002Y21} for central node while HGAN can address it by reducing the neighboring nodes through the hierarchical design.

% \begin{figure*}[!ht]
% \centering
% \includegraphics[width=0.9\linewidth]{figure/epoch.pdf}
% \caption{
% % For sentence embedding $v^{sen}$, we calculate the cosine value with token embedding from BERT $v^{token}$ and graph embedding $v^{graph}$ based on the 10k samples in ZESHEL test set.
% % We visualize the density distribution during the training process.
% We compare two fine-grained information: 1) learned from HGAN $v^{graph}$ 2) mean pooling of token embeddings of mentions/entities $v^{token}$ from BERT by calculating the cosine similarity with sentence embedding $v^{sen}$ on the ZESHEL test set. 
% The token embeddings $v^{token}$ are highly similar to the sentence embeddings $v^{sen}$, while the graph embeddings $v^{graph}$ keep  orthogonal. 
% }
% \label{epoch}
% \vspace{0.0em}
% \end{figure*}

\begin{itemize}
\item[\textbf{RQ3:}]
Can we add HGAN to mention or entity encoder only ?
\end{itemize}

\begin{table}[h]
\resizebox{\linewidth}{!}
{
\begin{tabular}{l|l|cccc}
\toprule
Mention Encoder & Entity Encoder & R@1   & R@8 & R@32  & R@64  \\ \midrule
BERT & BERT & 38.01 & 69.19 & 80.03 & 83.98 \\ 
BERT+HGAN & BERT & 38.16 & 69.41 & 80.04 & 83.92 \\ 
BERT & BERT+HGAN & 39.18 & 68.56 & 78.70 & 82.65 \\ 
BERT+HGAN & BERT+HGAN & 42.86 & 73.00 & 82.15 & 85.65 \\ 
\bottomrule
\end{tabular}
 }
\caption{
The ablation study results of the dual-encoder architecture.
(BERT, BERT) is the baseline BLINK while (BERT+HGAN, BERT+HGAN) is our proposed GER.
}
\label{ablation_study2}
\vspace{-2.0em}
\end{table}

% As shown in Figure \ref{framework}, 
GER follows the traditional dual-encoder framework where the mention encoder and entity encoder share the same architecture.
We conduct ablation studies that add the HGAN to mention encoder or entity encoder only.
Table \ref{ablation_study2} indicates that fusing fine-grained information for the encoder on one side will bring slight drops compared to the baseline BLINK.
Meanwhile, the GER framework performs better than BLINK.
For zero-shot entity retrieval, the mentions and entities should be embedded in one semantic space to match, thus fusing fine-grained information at one encoder leads to a drop in performance.

\subsection{Overall Performance}

% To evaluate the model performance under low resource scenarios, we downsample the size of the train set into 10\%, 20\%, 40\%, 60\% and 80\% while other parameters keep the same.
To evaluate the overall performance, we employ the BERT-base based cross-encoder for entity ranking after retrieving 64 candidate entities.
%All the parameters are kept the same.
For the first entity retrieval stage, it takes 48 minutes for our GER to train for 5 epochs on the train set of ZESHEL, and 40 minutes for BLINK. 

\begin{figure}[h]
	\centering
	\includegraphics[width=0.95\linewidth]{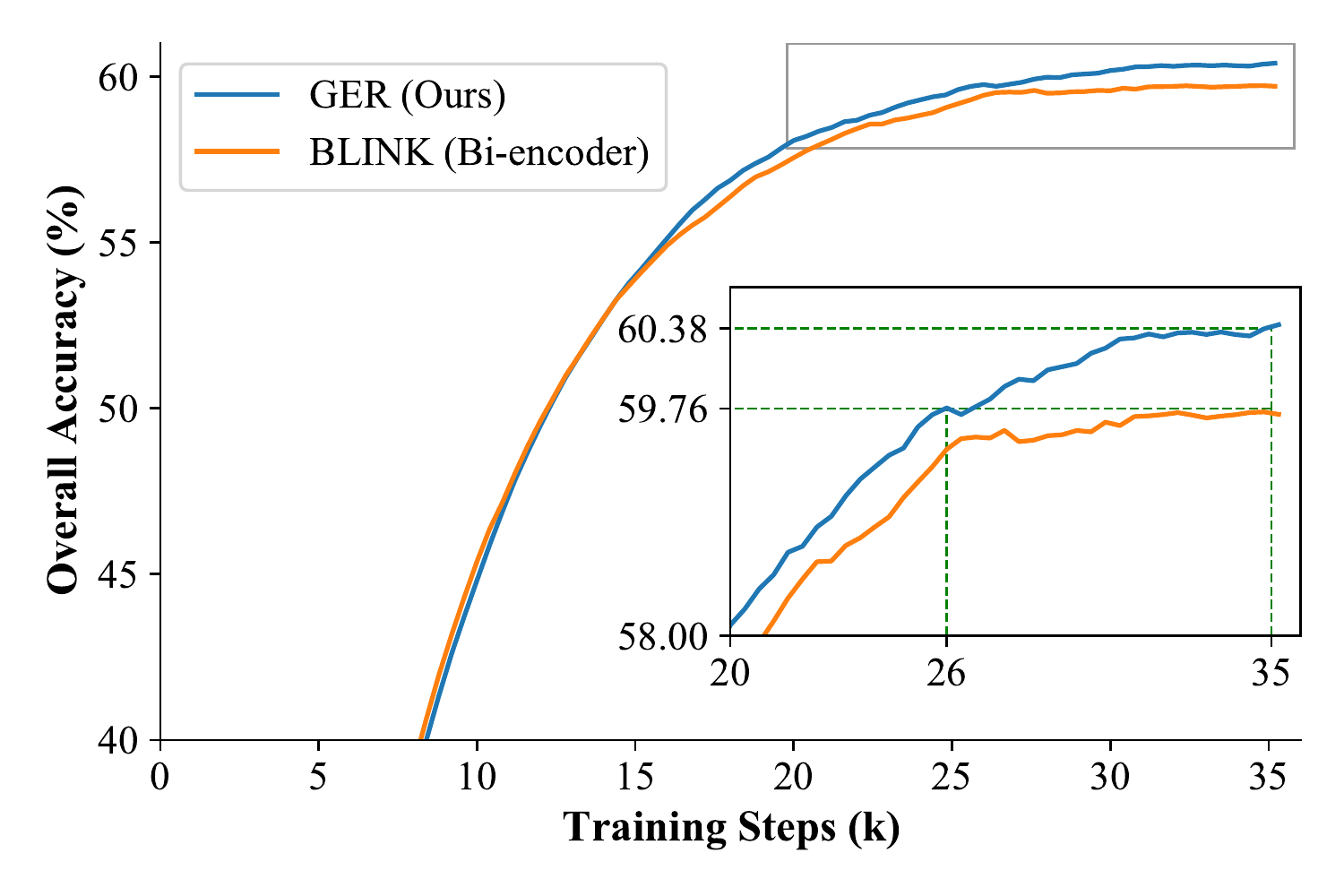}
	\caption{
	    Comparison of overall accuracy for BLINK and GER. 
	   % We employ BERT-base based cross-encoder for entity ranking stage.
	}
	\label{cross}
 	\vspace{-1.5 em}
\end{figure}

Figure \ref{cross} presents the overall accuracy of our GER and baseline BLINK on the test set of ZESHEL.
GER gets recall@64 of 85.65 compared to 83.98 from BLINK, and achieves 0.62 improvement after training 35K steps in entity ranking stage.
Moreover, GER get a comparable result in 26K steps with baseline BLINK in 35K steps, saving 9k steps~(nearly 2 hours).
In conclusion, GER achieves a higher recall@64 than BLINK during entity retrieval stage with slight extra 8 minutes, and such higher recall@64 can save 2 hours in the entity ranking stage.
% Figure \ref{few_sample} presents the performance of BLINK and GER on the test set of ZESHEL against different training settings.
% It shows that our framework GER achieves higher recall@64 score under all settings.
% With 60\% training samples, our GER achieves 85.05 score, 2.14 higher than BLINK. 
% Particularly, under the setting of 40\% training samples, our GER obtains better result~(84.01) than the BLINK model~(83.98) trained by all samples, indicating the effectiveness of our GER framework on mentions/entities modeling.

\begin{figure*}[!ht]
	\centering
	\includegraphics[width=0.95\linewidth]{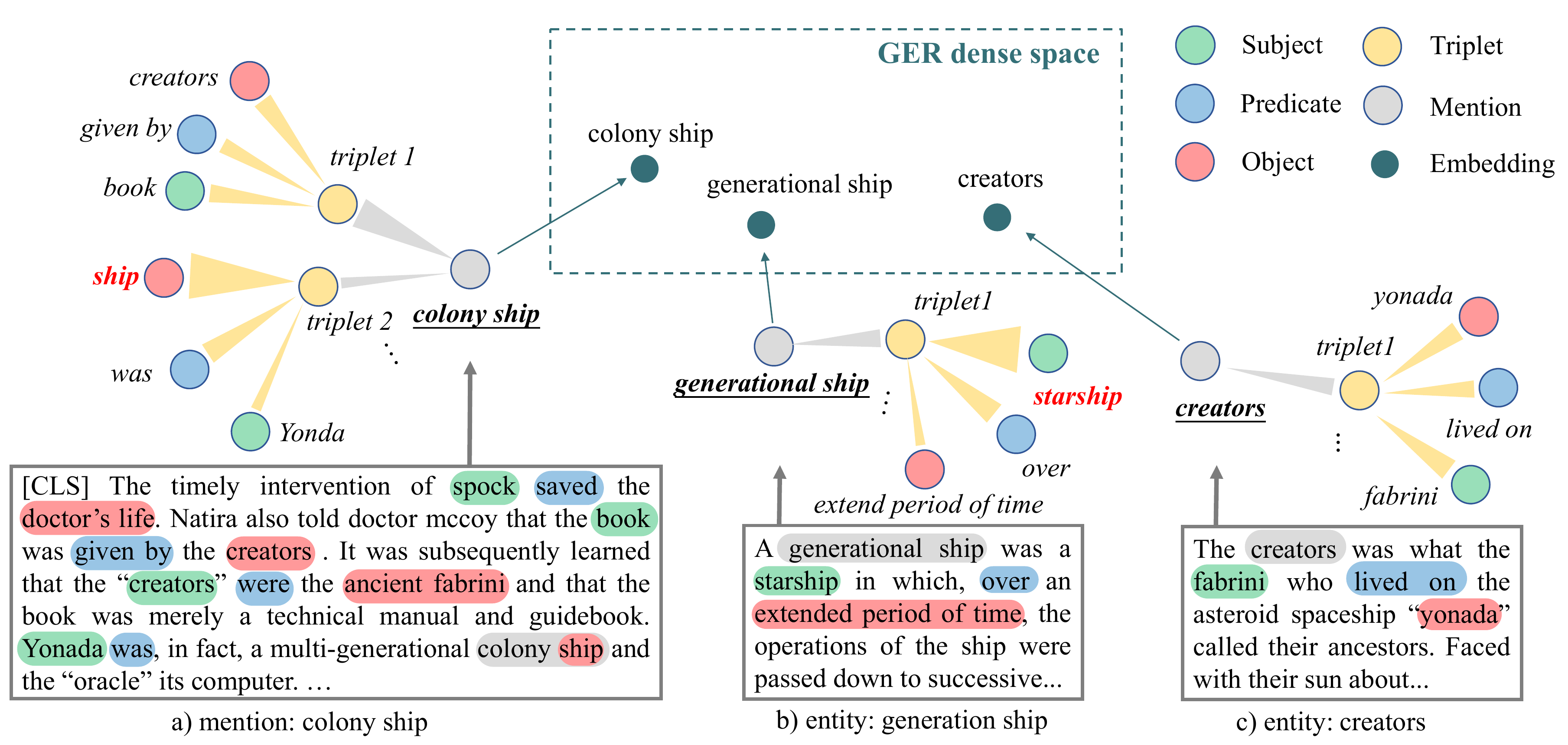}
	\caption{
		The corresponding embedding in dense space and part of node attention between graph nodes for mention \textit{colony ship}, ground truth entity \textit{generation ship} and entity \textit{creators}. 
		In the graph, we visualize the attention of Mention/Entity-Triplet edges (in grey) and Triplet-SPO edges (in yellow), where \textit{thicker edges} mean \textit{higher} attention scores. 
	}
	\label{case_study}
	\vspace{-1em}
\end{figure*}

\section{Attention Study}
% In this part, we compare the attention distribution for BLINK and GER.
% In this part, we explore the GER framework further by case study.
% Specifically, we compare the attention distribution in BERT for both baseline BLINK and our GER.
% Moreover, we also visualize the attention distributions in graph to reveal the knowledge learned by HGAN. 

\subsection{BERT Attention}

To investigate the attention distribution of the mention/entity token in BERT, we divide the 10000 samples in the ZESHEL test set into four groups by the rank of mention token attention.
For each sample, we calculate the attention scores between each token and \texttt{[CLS]} token in BERT and rank these tokens.
Then we group these samples by the lowest ranking of corresponding mention tokens.

\begin{table}[h]
\begin{tabular}{ll|cccc}
\toprule
\multicolumn{2}{l|}{Attention Ranking} & [0,32) & [32,64) & [64,96) & [96,128) \\
\midrule
\multirow{2}{*}{BLINK} & Total     & 685    & 1191    & 2765    & 5359     \\ 
                       & Recall@64 & 86.13  & 85.81   & 84.45   & 83.50    \\
\midrule
\multirow{2}{*}{GER}   & Total     & 742    & 1315    & 2798    & 5145     \\ 
                       & Recall@64 & 86.12  & 86.08   & 
                       86.78   & 84.98 \\
\bottomrule
\end{tabular}
\caption{
Attention distributions for ZESHEL test set.
% We group the 10000 samples in the ZESHEL test set by the attention ranking of mention among the all 128 input tokens in BERT, then compare the recall@64 for BLINK and our GER.
% The lower attention ranking, the higher attention scores to the \texttt{[CLS]} token.
}
\label{atten_distri}
\vspace{-1em}
\end{table}

As shown in Table \ref{atten_distri}, we can get the following observations:

(1) For BLINK, the higher the attention scores, the higher the recall performance. 
For the mentions ranked in [0,32), we get a recall@64 of 86.13, 2.63 higher than mentions ranked in [96,128).
The sentence embedding contains more information from the mention/entity when attention scores are higher.

(2) In our proposed GER, the attention scores towards mentions get higher, thus the sentence embeddings can model more information about the mentions.
For baseline BLINK, 53.59\% samples gain relatively low attention, where the attention ranking is [96,128).
For GER, we get a 2.14\% drop in [96,128), and these samples gain higher attention.
The sentence embeddings in GER can better model mentions/entities compared to baseline BLINK.
Since the node embeddings of knowledge-aware hierarchical graph are initialized by the token embedding, we believe that the knowledge in graph can influence the BERT layer through loss back propagation, thus the attention scores towards mention tokens gain.

\subsection{HGAN Attention}

To evaluate the ability to capture the fine-grained word-level information of mention/entity, we visualize the node-to-node attention weights in HGAN.
% , including the attention between the mention/entity node and triplet nodes and attention between triplets and corresponding SPO nodes. 
We calculate the edge attention following Equation \ref{atten_equation} and get the average among the attention heads.

As shown in Figure \ref{case_study}, for the embedding of mention \textbf{colony ship}, the embedding of entity \textbf{generation ship} is closer than entity \textbf{creators} in the dense space defined by GER.
% Compared to Figure \ref{example}, our GER captures fine-grained information as complementary to coarse-grained sentence embedding, thus model the relations between the mention and entity better.
In the knowledge-aware graph of mention \textbf{colony ship}, the node \textit{ship} in red gets the highest attention while the node \textit{creators} and \textit{yonda} receive relatively low attention, indicating that the mention encoder in GER focus on \textit{ship} and ignores \textit{creators}.
Meanwhile, for the entity \textbf{generational ship}, the node \textit{star ship} gets the highest score.
Through the knowledge-aware graph, HGAN gathers information from mention/entity related words to capture fine-grained information, and thus gets more comprehensive representations.
% Through the knowledge-aware graph, GER can model fine-grained information from mentions/entities and other related words, thus getting more comprehensive representations. 

\section{Conclusion}
% We propose GER, including graph enhanced encoder and symmetric loss, for zero-shot entity retrieval.
% Given the context, we extract the SPO triplets, thus building a mention/entity central graph, followed by a Hierarchy Graph Neural Network to capture the word information. 
% Our framework GER shows consistent improvements for the scale knowledge base ranging from 10K to 5.9M through the experiments. 
% Moreover, our framework performs better than the baseline model BLINK under few sample settings.
% For future work, we will exploit a better way to extract SPO triplets and apply the graph for candidate ranking.

% In this paper, we explore an issue that was overlooked when using BERT to retrieve entities in zero-shot entity linking scenarios, that is, high-frequency words dominate the sentence embedding, resulting in poor performance when the mention/entity a low-frequency word.
In this paper, we explore an issue that was overlooked when utilizing the sentence embedding from BERT to retrieve entities in zero-shot entity retrieval scenarios, that is, such coarse information can not fully model the mentions/entities and more fine-grained information is necessary.
Meanwhile, the token output are highly similar to the sentence embedding due to the over-smoothing in BERT.
To overcome these issues, we propose GER, which captures more fine-grained information from both mention/entity words and other related words as complementary to coarse-grained sentence embedding.
% Particularly, we extract the knowledge units from corresponding context and capture fine-grained information by a new designed Hierarchical Graph Attention Network~(HGAN).
Specifically, we construct a hierarchical graph and design a novel Hierarchical Graph Attention Network~(HGAN).
Experiments on several benchmarks show the effectiveness of our GER framework.
% Finally, we evaluate our approach on several popular zero-shot entity linking datasets and obtain a significant improvement compared with previous BERT-based models. 
Further analysis also indicate the importance of hierarchical design and symmetry design in GER.
In the future, we will consider applying the graph enhanced encoder for candidate ranking stage in zero-shot entity linking.

\section{Acknowledgments}
This work was partly supported by the National Key Research and Development Program of China~(No. 2020YFB1708200) and the Shenzhen Key Laboratory of Marine IntelliSense and Computation under Contract ZDSYS20200811142605016.

\clearpage
\normalem

\bibliographystyle{ACM-Reference-Format}
\bibliography{main}

\end{document}